\documentclass[conference]{IEEEtran}
\UseRawInputEncoding
\usepackage{cite}
\usepackage{amsmath,amssymb,amsfonts}
\usepackage{algorithmic}
\usepackage{graphicx}
\usepackage{textcomp}
\usepackage{xcolor}
\usepackage{float}
\usepackage{hyperref}
\usepackage{subfig}
\usepackage{multirow}
\usepackage{caption}
\usepackage{tablefootnote}

\begin{document}
\title{A Deep Learning Ensemble Framework for Off-Nadir Geocentric Pose Prediction}
\author{\IEEEauthorblockN{Christopher Sun}
\IEEEauthorblockA{\textit{Monta Vista High School} \\
Cupertino, CA, United States}
\and
\IEEEauthorblockN{Jai Sharma}
\IEEEauthorblockA{\textit{Monta Vista High School} \\
Cupertino, CA, United States}
\and
\IEEEauthorblockN{Milind Maiti}
\IEEEauthorblockA{\textit{Monta Vista High School} \\
Cupertino, CA, United States}}

\maketitle
\begin{abstract}
Computational methods to accelerate natural disaster response include change detection, map alignment, and vision-aided navigation. Current software functions optimally only on near-nadir images, though off-nadir images are often the first sources of information following a natural disaster. The use of off-nadir images for the aforementioned tasks requires the computation of geocentric pose, which is an aerial vehicle's spatial orientation with respect to gravity. This study proposes a deep learning ensemble framework to predict geocentric pose using 5,923 near-nadir and off-nadir RGB satellite images of cities worldwide. First, a U-Net Fully Convolutional Neural Network predicts the pixel-wise above-ground elevation mask of the RGB images. Then, the elevation masks are concatenated with the RGB images to form four-channel inputs fed into a second convolutional model, which predicts orientation angle and magnification scale. A performance accuracy of $\mathbf{R^2=0.917}$ significantly outperforms previous methodologies. In addition, outlier removal is performed through supervised interpolation, and a sensitivity analysis of elevation masks is conducted to gauge the usefulness of data features, motivating future avenues of feature engineering. The high-accuracy software built in this study contributes to mapping and navigation procedures for effective disaster response to save lives.
\end{abstract}


\section{Introduction}
An increasing frequency of natural disasters paralleled by a rapid increase in the availability of large satellite imagery data sets has introduced the challenge of big data analysis for practical disaster relief applications \cite{climate} \cite{thesis}. For an unmanned aerial vehicle (UAV) to effectively respond to natural disasters, versatile software must be able to perform a wide array of functionalities, such as detecting objects, monitoring changes on land, and mapping surroundings to allow for communication between the UAV and humans. While computer vision researchers have implemented algorithms to assist in search and rescue operations, there has been less progress in the tasks of vision-aided navigation and map alignment for disaster relief \cite{mishra}. These object detection algorithms perform most optimally on near-nadir images, where object parallax and occlusion are minimized. Oblique, off-nadir images can be repurposed for these algorithms through a semantic, three-dimensional knowledge of surroundings, but this information, known as geocentric pose, is unavailable when images are initially captured \cite{christie2021}. 

Geocentric pose consists of three components: above-ground elevation, angle of orientation with respect to gravity, and scale of magnification between the actual and apparent sizes of ground-level objects. The following two figures, adapted from the website of the Overhead Geopose Challenge hosted by the National Geospatial-Intelligence Agency, illustrate the concepts of scale and angle \cite{NGA}.
\begin{figure}[H]
    \centering
    \includegraphics[scale=0.38]{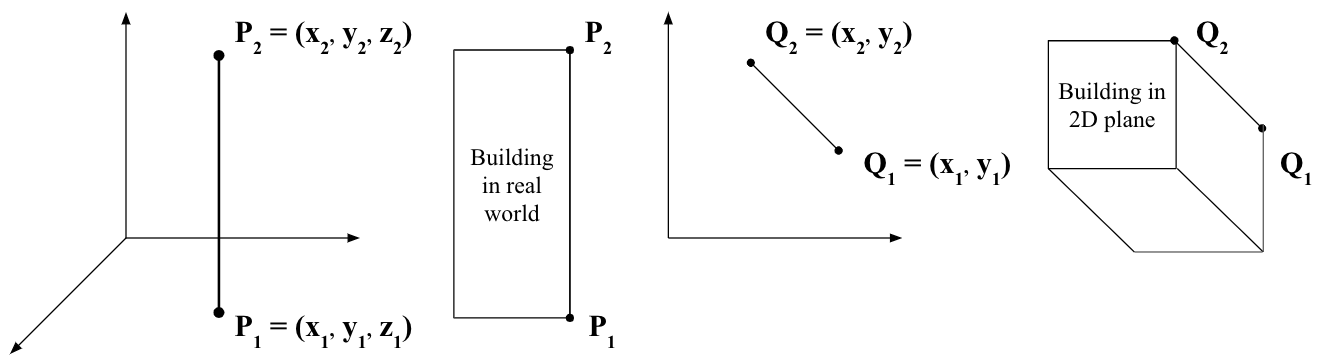}
    \caption{The scale in geocentric pose represents the magnification between building heights as represented on ground versus in a 2-D image plane.}
\end{figure}
\noindent Using the definitions in Fig. 1, $$\textrm{scale} = \frac{\lvert\lvert Q_2-Q_1 \rvert\rvert}{z_2-z_1}.$$ 
\begin{figure}[H]
    \centering
    \includegraphics[scale=0.4]{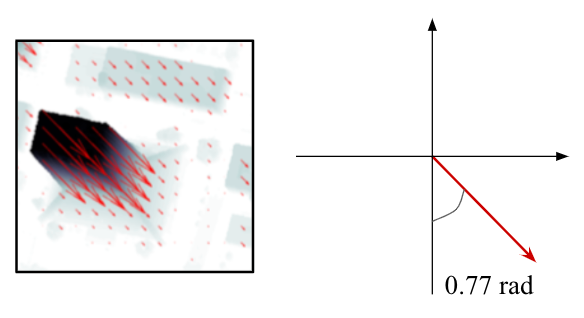}
    \caption{The angle in geocentric pose represents the degree of rotation between the y-axis and the flow vector field as determined by the direction of steepest vertical descent. The angle describes an object's orientation with respect to gravity.}
\end{figure}
Geocentric pose is harder to extract from off-nadir images compared to their near-nadir counterparts; yet, off-nadir images are the most immediate sources of information following natural disasters \cite{offnadir}. This limitation creates the need for an efficient and accurate computational methodology for off-nadir geocentric pose prediction.

There has been limited research on deep-learning-aided geocentric pose prediction from satellite imagery. One of the few works that used this approach is that of Christie et al., who created an encoding for geocentric pose using monocular satellite imagery, attempting to rectify object parallax of monocular images and thereby improving the accuracy of object localization for Earth observation tasks \cite{christie2020} \cite{christie2021}. The main shortcoming of this work was the accuracy of geocentric pose prediction.

This paper outlines a novel ensemble methodology to predict the geocentric pose of near-nadir and oblique RGB satellite images. First, a fully convolutional U-Net model is used to predict the pixel-wise elevations of RGB satellite images. Then, a convolutional model with subsequent fully connected layers is used to predict the scale of magnification and angle of orientation, given the RGB image as well as the predicted pixel-wise elevations. 
\begin{figure}[H] 
    \centering
    \includegraphics[scale=0.45]{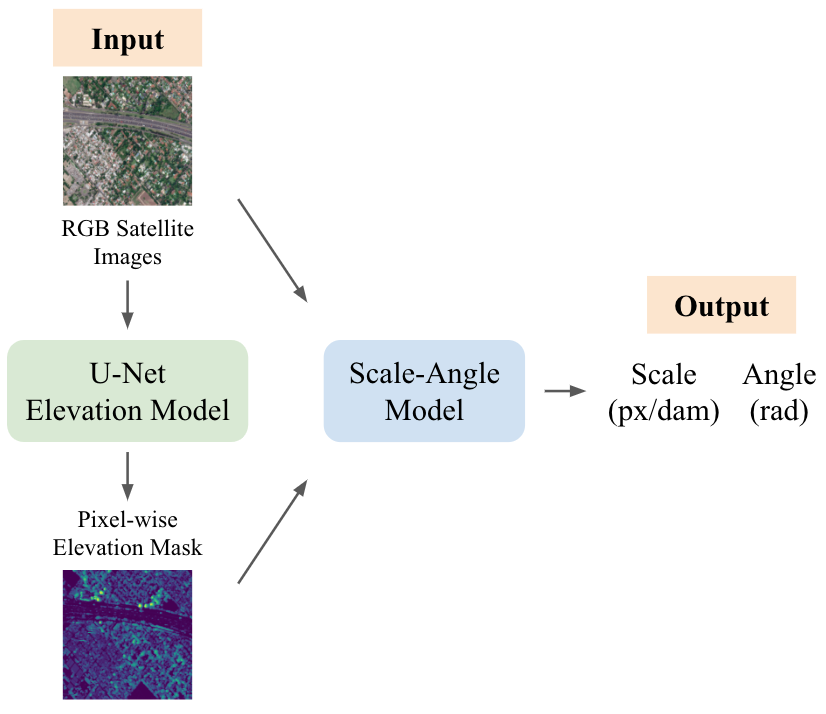}
    \caption{The flowchart above depicts the ensemble framework to predict geocentric pose.}
\end{figure}

\section{Materials and Methods}
\subsection{Data Used For Experimentation}
The data used for this research is part of the Urban Semantic 3D Data Set, publicly available from the IEEE DataPort \cite{data}. The data set contains 5,923 RGB images, their pixel-wise above-ground elevations\footnote{Some elevation masks contain ``Not a Number'' (NaN) pixel values. The solution to this problem is discussed in Section \ref{sec:interpolation}.}, and their measurements of scale and angle. The images were captured from four general locations in North and South America.
\begin{table}[H]
    \centering
    \renewcommand{\arraystretch}{1.2}
    \caption{Geographical Sources of Data}
    \begin{tabular}{|c|c|}
    \hline
    \textbf{Location} & \textbf{No. of Images} \\ \hline\hline
    San Fernando, AR  & 2325                   \\ \hline
    Atlanta, USA      & 704                    \\ \hline
    Jacksonville, USA & 1098                   \\ \hline
    Omaha, USA        & 1796                   \\ \hline
    \end{tabular}
\end{table}
Elevation was given in centimeters, scale was given in pixels per centimeter\footnote{The unit for scale was converted to pixels per decameter when reporting metrics.}, and angle was given in radians. Due to computational constraints, image dimensions were resized from $2048\times2048\times3$ to $256\times256\times3$ using interpolation through resampling by pixel area relation.
\subsection{Outlier Removal}
The data set contained some elevation masks with improbably large pixel values, higher than the tallest building in the city the data was acquired from. The solution to this problem was location-based thresholding, which involved finding pixel value thresholds for each city in Table 1. Images from Jacksonville and Omaha did not contain any outliers. For each image from Atlanta, the number of pixels $N$ that exceeded 4,000 centimeters was counted. If $N$ exceeded 100, we determined that the image indeed contained a building taller than 4,000 centimeters. If $N$ was less than 100, all said pixels were interpolated with the median elevation in the image. For images from San Fernando, the outlier threshold was a hard cap of 3,000 centimeters, because the landscape was fairly even and did not contain tall buildings. 

One limitation of location-based thresholding was that thresholds were experimentally determined, since there was no objective way to best deal with outliers. In particular, the challenge was discriminating between large pixel values that were indeed invalid and those that were valid due of the presence of a tall building. 
\subsection{Autoencoder Model}
Before embarking on the ensemble framework, an autoencoder was trained on the RGB images to gauge the practicality of a deep learning approach. Our reasoning was that if meaningful latent space representations could be produced from the RGB images, the other convolutional models would likewise have the capability to extract useful features to predict geocentric pose. 
\begin{figure}[H]
    \centering
    \includegraphics[scale=0.25]{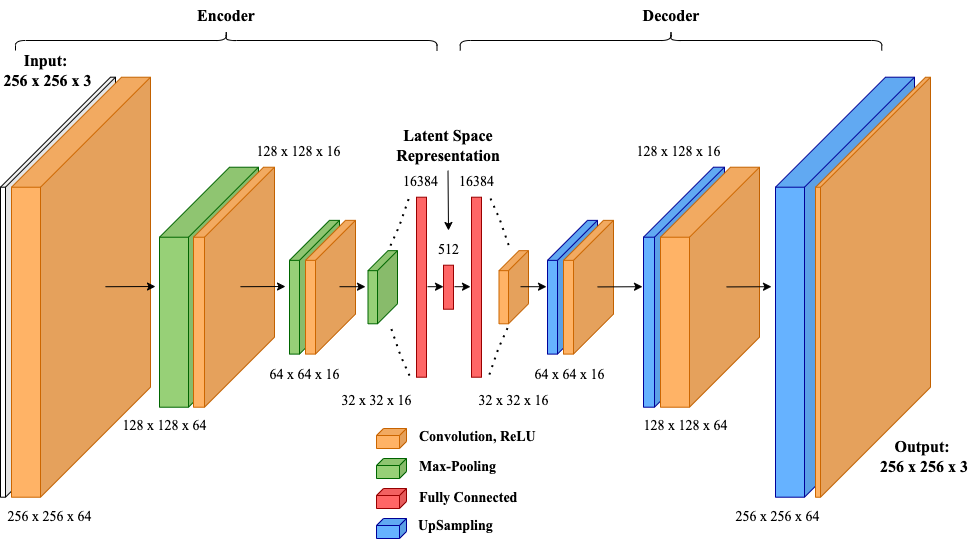}
    \caption{Shown above is the architecture of the Autoencoder Model trained with 16,823,107 parameters, consisting of an encoder (left) and a decoder (right).}
\end{figure}
The encoder used convolutional operations to condense the input images to intermediate matrices of shape $32\times 32\times 16$. These matrices were then flattened to a vector of length $16384$, after which a dense layer was used to arrive at the latent space representation of length $512$. Multidimensional Scaling (MDS) was used to visualize these high-dimensional latent space representations on a two-dimensional plane. An inverse architecture of the encoder was used to decode the latent space representations back into images of shape $256 \times 256\times 3$. The autoencoder, a result of combining the encoder and decoder into one model, was trained with the goal of retrieving decoded images as similar to the original input images as possible. The accomplishment of this task would confirm the feasibility of the extraction of high-level features using computer vision techniques.

\subsection{U-Net Elevation Model}
\begin{figure}[H]
    \centering
    \includegraphics[scale=0.25]{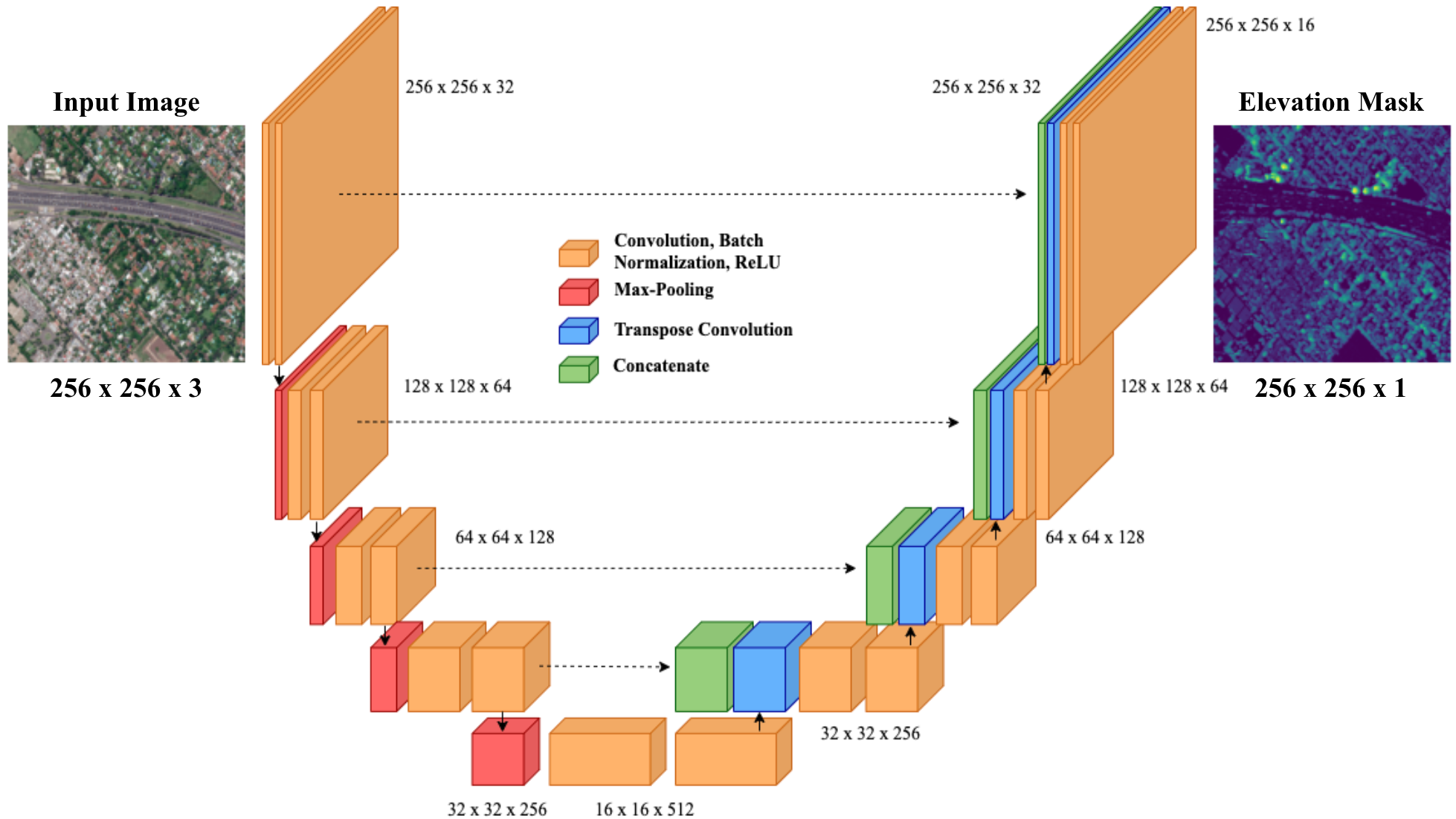}
    \caption{Shown above is the architecture of the U-Net Elevation Model trained with 23,966,337 parameters. Skip connections between downsampling and upsampling blocks allow for the concatenation of past feature information during later stages of processing.}
\end{figure}
A fully convolutional neural network, the first model in the ensemble framework, was trained to predict the pixel-wise elevation mask of each RGB image. The model was based on a U-Net architecture designed for a wildfire detection task, with modifications such as the addition of extra convolutional blocks and skip connections between downsampling and upsampling layers \cite{wildfire}. The filter size was $5 \times 5$ throughout the model. Starting with 32 filters, the number of filters was doubled for each layer during downsampling and halved for each layer during upsampling. Batch Normalization was employed between convolution layers for a regularization effect, and the ReLU activation function was used throughout the model. The model was configured with an 85-15\% train-validation split and trained until convergence using Adam optimization, a constant learning rate of 0.001, and a mini-batch size of 32 images \cite{adam}. Images with ``Not a Number'' (NaN) values were withheld from the training data set for data cleaning purposes described in Section \ref{sec:interpolation}.

The described model will henceforth be referred to as the U-Net Elevation Model.

\subsection{Scale-Angle Model} \label{sec:geopose methods}
\begin{figure}[H]
    \centering
    \hspace*{-1mm}\includegraphics[scale=0.17]{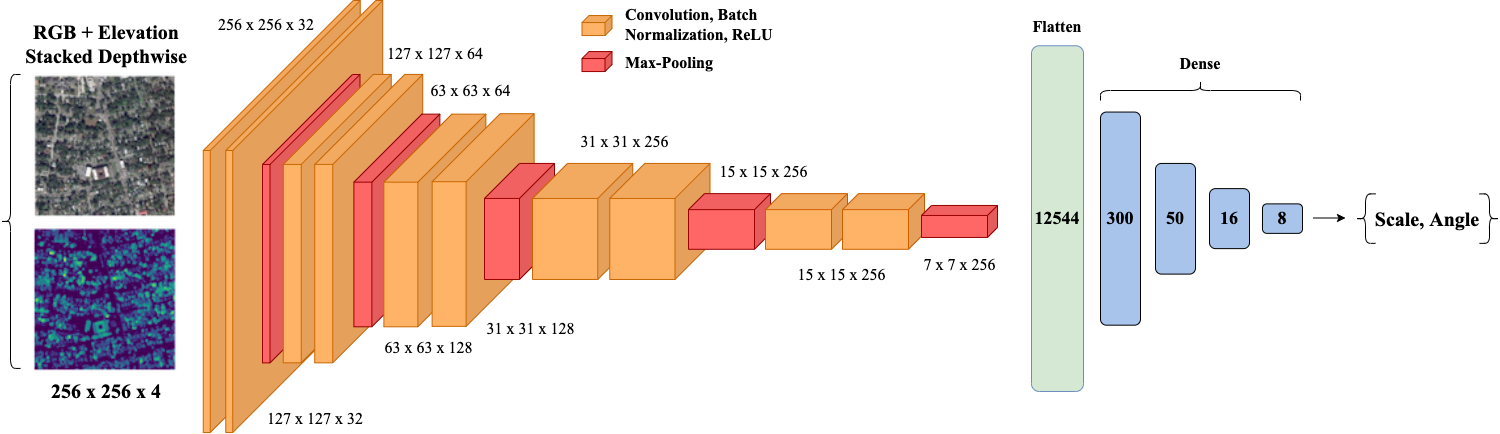}
    \caption{Shown above is the architecture of the Scale-Angle Model trained with 7,076,370 parameters. RGB images and their elevation masks were stacked depthwise and fed through the model, which predicted the remaining two components of geocentric pose.}
\end{figure}
The second model in the ensemble framework, referred to as the Scale-Angle Model, was a convolutional neural network with a series of five double-convolution layers and maximum pooling in between each series. The resulting $7 \times 7 \times 256$ shape was flattened and fed through six fully connected layers\footnote{Fig. 6 only displays five fully connected layers. The sixth layer contained 2 hidden nodes, corresponding to scale and angle, respectively.} to generate the scale and angle predictions. The model was configured with an 80-20\% train-validation split and trained until convergence using the same specifications as the U-Net Elevation Model.

After elevation mask predictions were obtained from the U-Net Elevation Model, they were concatenated with the original RGB images along the channels axis to produce four-channel images. These images were then fed as inputs to the Scale-Angle Model for the prediction of scale and angle. Together, the U-Net Elevation Model and the Scale-Angle Model predicted all three components of geocentric pose.

\section{Results}
\subsection{Autoencoder: Latent Space Representations\protect\footnote{It must be noted that these visualizations do not capture all nuances of the high-dimensional data, so they are used as proof of concept.}}
\begin{figure}[H]
    \centering
    \hspace*{-2mm}\includegraphics[scale=0.45]{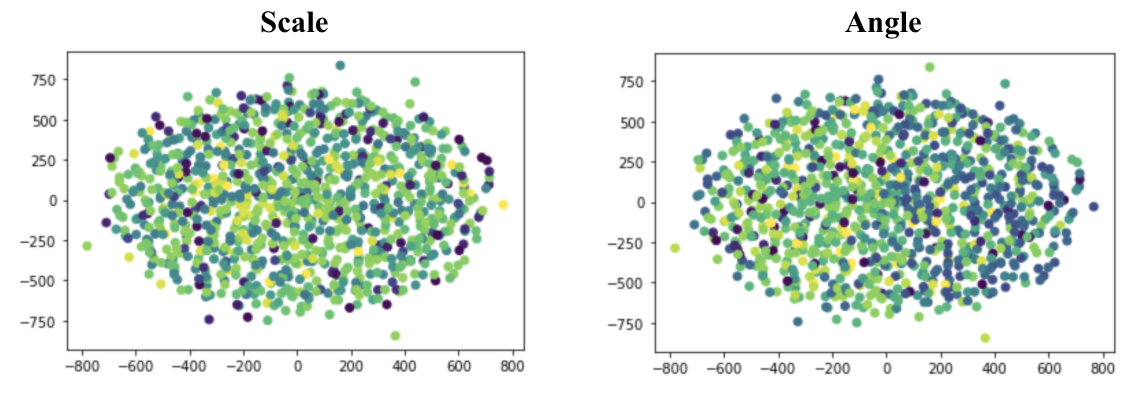}
    \caption{MDS was employed to visualize the latent spaces representations of the encoded images. In the figures above, each point represents an encoded image, which is colored by the magnitude of its scale and angle, respectively.}
\end{figure}
\paragraph{Visualization for Scale}
No clear association was observed from the visualization of scale encodings. This may have been caused by the decrease in variance due to dimensionality reduction.
\paragraph{Visualization for Angle}
Contrary to the results of the scale encodings, the visualizations for angle encodings showed a clearer association. Namely, images with higher angles were aggregated spatially to the left while images with lower angles were aggregated spatially to the right. The significance of this result was twofold, demonstrating not only that there is a relationship between an RGB image and its angle, but also that this said relationship can be learned through a convolutional approach. Hence, we considered it feasible to use deep learning techniques to learn this relationship and predict the geocentric pose of RGB images.

\subsection*{Evaluation Metrics}
In addition to the mean absolute error (MAE) and median absolute deviation (MAD) metrics, the models in the ensemble framework were evaluated by the $R^2$ metric, defined as $$R^2=1-\frac{\sum_{i=1}^{n}(y_i-\hat{y_i})^2}{\sum_{i=1}^{n}(y_i-\bar{y})^2},$$ where $n$ is the number of values, $y_i$ is the $i$th true value, $\hat{y_i}$ is the $i$th predicted value, and $\bar{y}$ is the mean of all true values.
\subsection{U-Net Elevation Model Performance}
The U-Net Elevation Model achieved an $R^2$ metric of 0.926 on training data and 0.865 on validation data. The model was prone to minimal overfitting, with a high degree of generalizability to unseen data, which was both numerically and visually confirmed.
\subsection*{Interpolation of Invalid Data} \label{sec:interpolation}
Recall that images containing NaN values were excluded from training. This was to allow NaN pixels to be replaced with the predictions of the U-Net Elevation Model after convergence on valid data. Interpolation was performed on all NaN values, serving as a data cleaning and quality control mechanism that allowed all 5,923 images to be used for the development of the Scale-Angle Model.
\begin{figure}[H]
    \centering
    \hspace*{-1mm}\includegraphics[scale=0.6]{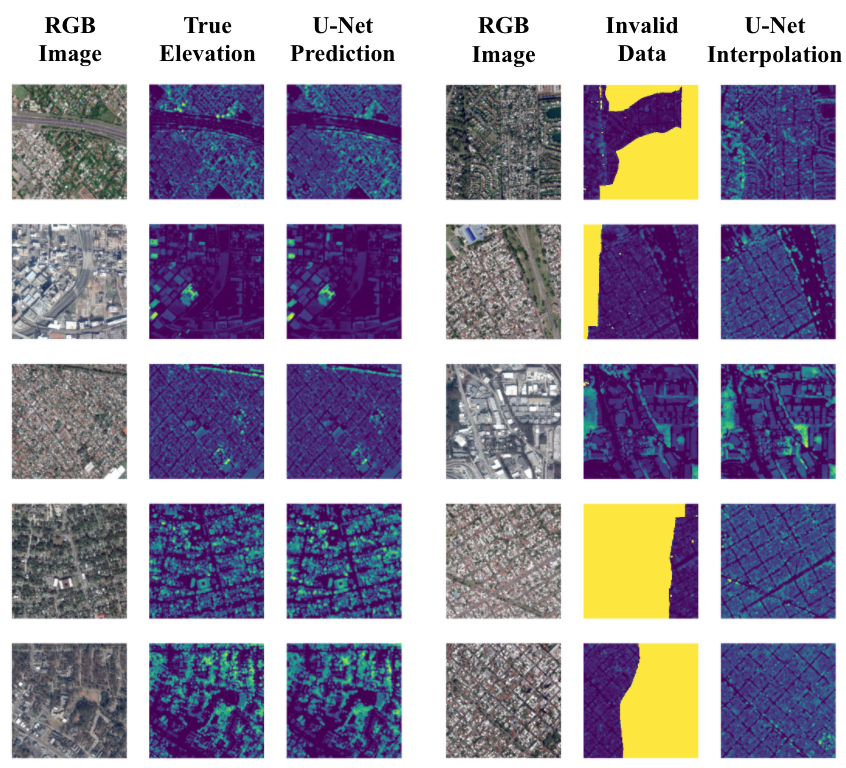}
    \caption{The elevation masks predicted by the model are compared to the true elevation masks (left). Notice the similarity between the true and predicted elevation masks. The model interpolations are displayed next to their counterparts containing large amounts of invalid pixel values (right). Notice how the third row contains a small yellow patch of NaN values near the top left of the elevation mask. Images like this required interpolation for only a small number of pixels. On the other hand, for images containing mostly NaN values, such as the fourth and fifth rows, most of the elevation mask was reconstructed using model predictions.}
\end{figure}
\subsection{Scale-Angle Model Performance}
\begin{figure}[H]
    \centering
    \hspace*{-1.5mm}\includegraphics[scale=0.24]{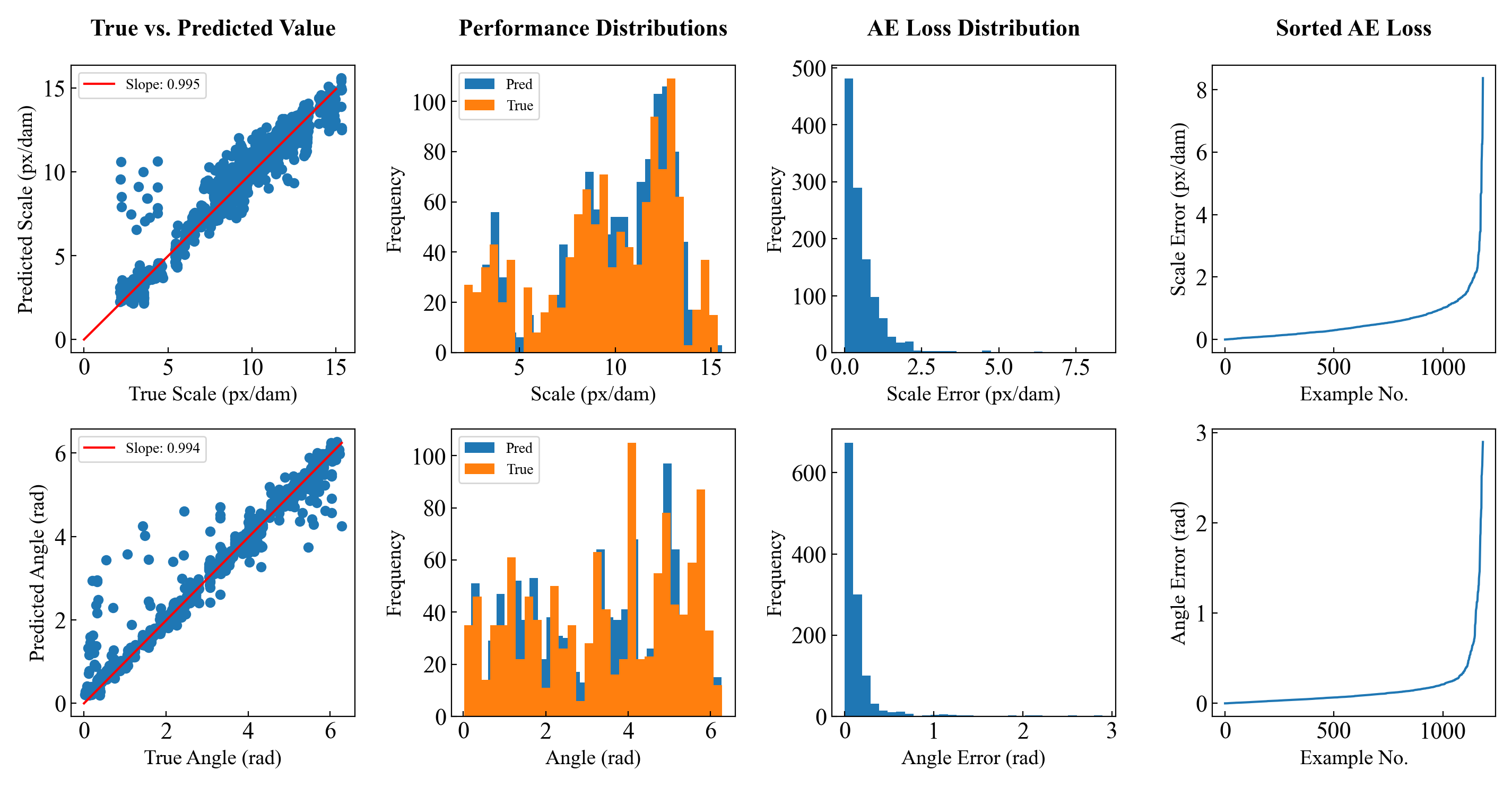}
    \caption{The above plots show the resemblance between true and predicted values as well as the distribution of error of the Scale-Angle Model.}
\end{figure}
The Scale-Angle Model achieved an $R^2$ metric of 0.991 on training data and 0.943 on validation data, weighed equally between scale and angle. The following conclusions are derived from Fig. 9.
\paragraph{True vs. Predicted Value}
The predicted scale and angle measurements for all images are plotted against the true scale and angle measurement, with linear regression coefficients of 0.995 and 0.994, respectively. 
\paragraph{Performance Distributions}
The distributions of the true and predicted values have similar shape and spread for both scale and angle, showing the model's ability to capture trends between the RGB-and-Elevation inputs and the corresponding geocentric pose.
\paragraph{Distribution of Absolute Error Loss}
The distributions of absolute error loss for both scale and angle are heavily skewed right, which shows that the model rarely outputs predictions that are faulty to a large degree. The distribution for angle is more skewed than that for scale.
\paragraph{Sorted Absolute Error}
The graphs of sorted absolute error for both scale and angle take a skewed shape. On average, the model was able to predict the scale to within 0.386 pixels per decameter, and the angle to within 0.079 radians ($\approx4.53^{\circ}$). 

The final metrics for the geocentric pose ensemble framework are summarized below.
\begin{table}[H]
    \centering
    \renewcommand{\arraystretch}{1.2}
    \caption{Geocentric Pose Ensemble Framework Metrics}
    \begin{tabular}{|p{0.6cm}||cc||cccc|}
    \hline
                 & \multicolumn{2}{c||}{\textbf{U-Net Model}}       & \multicolumn{4}{c|}{\textbf{Scale-Angle Model}}                                                                                        \\ \hline\hline
                 & \multicolumn{2}{c||}{\textbf{Elevation} (m)}                   & \multicolumn{2}{c|}{\textbf{Scale} (px/dam)}                                            & \multicolumn{2}{c|}{\textbf{Angle} (rad)}                       \\ \hline
    \centering Metric       & \multicolumn{1}{p{0.8cm}|}{\centering\textit{Train}} & \multicolumn{1}{p{0.8cm}||}{\centering\textit{Val}} & \multicolumn{1}{p{0.8cm}|}{\centering\textit{Train}} & \multicolumn{1}{p{0.8cm}|}{\centering\textit{Val}} & \multicolumn{1}{p{0.8cm}|}{\centering\textit{Train}} & \multicolumn{1}{p{0.8cm}|}{\centering\textit{Val}} \\ \hline
    \centering\textbf{MAD} & \multicolumn{1}{c|}{0.263}           & 0.335                & \multicolumn{1}{c|}{0.192}          & \multicolumn{1}{c|}{0.386}               & \multicolumn{1}{c|}{0.063}          & 0.079               \\ \hline
    \centering\textbf{MAE} & \multicolumn{1}{c|}{0.847}          & 1.049               & \multicolumn{1}{c|}{0.243}          & \multicolumn{1}{c|}{0.582}               & \multicolumn{1}{c|}{0.091}          & 0.160               \\ \hline
    \centering$\mathbf{R^2}$  & \multicolumn{1}{c|}{0.926}          & 0.865               & \multicolumn{1}{c|}{0.990}          & \multicolumn{1}{c|}{0.924}               & \multicolumn{1}{c|}{0.991}          & 0.963               \\ \hline
    \end{tabular}
\end{table}
\section{Discussion}
\subsection{Insights from Deep Learning Models}
\paragraph{Autoencoder Model}
The latent space representations of the RGB images were twelve times smaller in storage compared to the original images, yet preserved important features within the data. The ability of the autoencoder to reproduce the RGB images motivates future avenue of exploration that involve using autoencoders as feature extractors that can feed high-level characteristics of RGB images into the ensemble framework.
\paragraph{Feature Importance for Ensemble Framework}
We sought to understand the usefulness of the features extracted by the U-Net Elevation Model by testing the degree of importance of the elevation masks in regards to geocentric pose prediction. This was accomplished through a sensitivity analysis involving three different versions of the Scale-Angle Model, as follows.
\begin{enumerate}
    \item \textbf{RGB-Only Model:} The RGB-Only Model received only the RGB images as inputs, acting as a negative control.
    \item \textbf{RGB-Elevation Model} The RGB-Elevation Model was the standard Scale-Angle Model.
    \item \textbf{Elevation-Only Model} The Elevation-Only Model received only the elevation masks as inputs, acting as a baseline performance standard.
\end{enumerate}
These three models were trained with the same hyperparameter specifications as described in Section \ref{sec:geopose methods}, with the exception of the input shape, which differed along the channels axis. The $R^2$ of each of the three defined models and the \textit{P}-values between adjacent $R^2$ metrics, calculated as a difference in proportions, is listed below, with the models numbered in the order they were introduced. 
\begin{table}[H]
    \centering
    \renewcommand{\arraystretch}{1.2}
    \caption{Scale-Angle Model Sensitivity Analysis}
    \begin{tabular}{|c|c|c|c|c|c|}
    \hline
                   & \textbf{1) $\mathbf{R^2}$} & \textbf{\textit{P}-value}        & \textbf{2) $\mathbf{R^2}$} & \textbf{\textit{P}-value}       & \textbf{3) $\mathbf{R^2}$} \\ \hline\hline
    \textbf{Scale} & 0.591    & 8.55E-80 & 0.924         & 2.24E-4 & 0.881          \\ \hline
    \textbf{Angle}    & 0.733    & 7.30E-55 & 0.963         & 5.16E-1\tablefootnote{The Elevation-Only Model outperformed the RGB-Elevation Model in angle predictions, an opposite manifestation of the alternative hypothesis, hence the \textit{P}-value $>$ 0.5.} & 0.963          \\ \hline
    \end{tabular}
\end{table}
The $R^2$ of the RGB-Elevation Model exceeded that of the RGB-Only Model for both scale and angle with a statistically significant \textit{P}-value. Similarly, the $R^2$ of the RGB-Elevation Model exceeded that of the Elevation-Only Model for scale with a statistically significant \textit{P}-value. The Elevation-Only Model surprisingly outperformed the RGB-Elevation Model in angle predictions, but by an extremely small margin.

The results of the sensitivity analysis show that the presence of elevation masks as inputs is integral for the Scale-Angle Model to achieve high performance metrics, meaning above-ground elevation is a crucial feature for the prediction of the scale and the angle. Elevation masks could specifically be beneficial in predicting the angle of orientation, as knowledge of relative building heights can assist in determining their orientation. On the other hand, processing RGB images in conjunction with elevation masks could provide useful information about the scale of magnification, which relies on a measure of the conversion factor between pixels in the image and distances in the real world. 
\subsection{Limitations and Future Improvements}
When analyzing specific predictions of low accuracy, an unexpected result was that the Scale-Angle Model struggled to predict the angle of some near-nadir images, likely because the height of buildings are difficult to perceive at these angles, rendering it difficult to distinguish between angles in this range.

As stated previously, computational constraints necessitated the downsizing the input images, which lowered images' resolutions. Instead, random cropping can be employed to accomplish the same task while preserving the original resolution. Performance can further be enhanced through image augmentation, particularly for the cities of Atlanta and Jacksonville, to prevent the model from overfitting to cities with the most labeled data. Additionally, the ensemble framework may benefit from the usage of the Ground Sample Distance (GSD) feature included in the meta-data.

Future research should explore optimal outlier removal methods for this data set. Further experimentation is also necessary to investigate how the model architectures described in this paper could be improved. For example, changing the size of the bottleneck layer in the U-Net Elevation Model or using two separate, simpler models to predict scale and angle might result in better performance. In addition, labeled geocentric pose data from contrasting landscapes is required to understand the ensemble's degree of generalization, since the ensemble was trained mainly on suburban and urban images.

\subsection{Comparison to Prior Research}
While Christie et al. achieved a cumulative $R^2$ metric of 0.799, the models described in this study achieved a cumulative $R^2$ metric of 0.917, albeit on different testing examples \cite{christie2021}. Nevertheless, this demonstrates a major increase in accuracy and reliability of geocentric pose predictions. We attribute this higher performance to the ensemble methodology as well as the data preprocessing and supervised interpolation that served as data quality control. 
\section{Conclusion}
The ensemble framework proposed in this paper will improve the versatility of natural disaster response software through the prediction of geocentric pose for both near-nadir and off-nadir aerial vehicles. Validated by an $R^2$ metric superior to previous research investigations, our results will allow for the repurposing of off-nadir images for disaster response tasks such as vision-aided navigation, change detection, and map alignment tasks for effective disaster response.

\section*{Additional Information}
The authors have made the code for the research in this paper open source and available on \href{https://github.com/jaisharmz/geocentric-pose}{GitHub (click)} \cite{code}.
\end{document}